\documentclass[a4paper,fleqn]{cas-dc}

\usepackage[numbers]{natbib}

\usepackage{amsmath,amsfonts}
\usepackage{algorithmic}
\usepackage{algorithm}
\usepackage{array}
\usepackage{graphicx}
\usepackage{booktabs}
\usepackage{multirow}
\usepackage{makecell}
\usepackage{multicol}

\newtheorem{theorem}{Theorem}[section]

\def\tsc#1{\csdef{#1}{\textsc{\lowercase{#1}}\xspace}}
\tsc{WGM}
\tsc{QE}
\tsc{EP}
\tsc{PMS}
\tsc{BEC}
\tsc{DE}

\begin{document}
\let\WriteBookmarks\relax
\def\floatpagepagefraction{1}
\def\textpagefraction{.001}
\shorttitle{Broad stochastic configuration residual learning system}
\shortauthors{Han Su et~al.}

\title [mode = title]{Broad stochastic configuration residual learning system for norm-convergent universal approximation}                      
\tnotemark[1]
\tnotetext[1]{This research is supported by National Key R\&D Program of China (No.2020YFB1707802), National Natural Science Foundation of China (No.12071131,No.11571107).}

\author[1]{Han Su}[orcid=0000-0002-1260-5782]
\ead{hdsuhan@ncepu.edu.cn}


\affiliation[1]{organization={School of Control and Computer Engineering, North China Electric Power University},
                city={Beijing},
                postcode={102206}, 
                country={China}}

\author[2]{Zhongyan Li}[orcid=0000-0002-0132-730X]
\cormark[1]
\ead{lzhongy@ncepu.edu.cn}


\affiliation[2]{organization={School of Mathematics and Physics, North China Electric Power University},
                city={Beijing},
                postcode={102206}, 
                country={China}}

\author[3]{Wanquan Liu}[orcid=0000-0003-4910-353X]
\cormark[1]
\ead{liuwq63@mail.sysu.edu.cn}

\affiliation[3]{organization={School of Intelligent Systems Engineering, Shenzhen Campus of Sun Yat-sen University},
                acity={Shenzhen},
                postcode={518107}, 
                country={China}}

\cortext[cor1]{Corresponding author}


\begin{abstract}
Universal approximation serves as the foundation of neural network learning algorithms.
However, some networks establish their universal approximation property by demonstrating that the iterative errors converge in probability measure rather than the more rigorous norm convergence, which makes the universal approximation property of randomized learning networks highly sensitive to random parameter selection.
Broad residual learning system (BRLS), as a member of randomized learning models, also encounters this issue. We theoretically demonstrate the limitation of its universal approximation property, that is, the iterative errors do not satisfy norm convergence if the selection of random parameters is inappropriate and the convergence rate meets certain conditions.
To address this issue, we propose the broad stochastic configuration residual learning system (BSCRLS) algorithm, which features a novel supervisory mechanism adaptively constraining the range settings of random parameters on the basis of BRLS framework. 
Furthermore, we prove the universal approximation theorem of BSCRLS based on the more stringent norm convergence.
Three versions of incremental BSCRLS algorithms are presented to satisfy the application requirements of various network updates.  
Solar panels dust detection experiments are performed on publicly available dataset and compared with 13 deep and broad learning algorithms. Experimental results reveal the effectiveness and superiority of BSCRLS algorithms.
\end{abstract}


\maketitle

\section{Introduction}

In recent years, neural networks have received widespread applications with outstanding performance in the field of machine learning \cite{DeepMachineLearning} owing to their universal approximation ability for nonlinear mappings \cite{NonlinearMaps} and their capacity to learn from training sample sets \cite{TraningSample}. 
The universal approximation property  \cite{UniversalApproximation} serves as the cornerstone of neural network learning algorithms, and enables a network to achieve a good balance between learning and generalization. 
For randomized learning algorithms, the input parameters are randomly assigned and the output weights are assessed using the least squares and regularization. 
Therefore, the universal approximation property of randomized learning algorithms is generally proven based on the iterative errors converging in probability measure rather than in more rigorous norm convergence. 
However, the universal approximation property with convergence in probability measure makes the network performance sensitive to random parameter selection, where improper settings of random parameters could result in the network failing to approximate the target function with a high probability.
This occurs because the convergence in probability measure is a less stringent condition that allows for the possibility that the algorithm's performance might not consistently reach excellent results across all possible outcomes. 
In essence, it holds true in a probabilistic sense over a large number of trials. 
In contrast, norm convergence requires that the iterative errors of the algorithm converge to zero in a more absolute and deterministic manner. 
In fact, norm convergence is often difficult to achieve for random learning algorithms where the input parameters are not fixed but randomly assigned. 
As a form of randomized learning algorithms, broad learning networks have the universal approximation property with convergence in probability measure \cite{BLS_Universal}, and also face the issue where network feasibility is influenced by parameter selection. 
To solve the parameter sensitivity issue, we focus on constructing a novel broad learning network that has the universal approximation property with iterative errors converging in norm.


Broad learning networks have a distinct advantage of reducing training time and maintaining learning capabilities, compared to deep learning networks with computationally expensive training process and complex hyperparameter tuning \cite{BLS1}.
Benefiting from its excellent performance, broad learning networks have been widely applied and made ground-breaking contributions to privacy protection \cite{BLS_PrivacyProtection}, label distribution learning \cite{BLS_LabeDistributionLearning}, data clustering \cite{BLS_DataClustering} and so on.
In practice, broad learning networks unavoidably exist the degradation issue, which stems from the indetermination of linear independence of incremental nodes driven by the parameter randomness \cite{BLS2}. The degradation issue could impact the universal approximation capability as the network scale grows larger. 
To tackle the issue, broad residual learning system (BRLS) \cite{BRLS} constructs a residual operator sequence with norm convergence through a novel incremental mechanism.
Although BRLS enhances the universal approximation capability of broad learning networks by improving degradation, BRLS does not address the parameter sensitivity issue caused by universal approximation property with convergence only in probability measure.
To overcome this issue, it is necessary to construct an improved BRLS network that can constrain random parameters to meet a more rigorous universal approximation property with convergence in norm.

To build the target network, we focus on the classic random vector functional link (RVFL) network \cite{RVFL}, which is the foundational theory for broad learning. RVFL has the same issues as BRLS, where the universal approximation property with converging in probability measure makes the feasibility of the network reliant on the random parameter selection. 
Researches indicate that the universal approximation property of RVFL is valid for random parameters selected from a uniform distribution within a defined range \cite{RVFL1}. 
The property extends to symmetric interval settings for random parameter range when the approximated function meets Lipschitz condition \cite{RVFL2}. 
Furthermore, studies both empirically \cite{RVFL3} and theoretically \cite{RVFL4} demonstrate the indetermination of RVFL's universal approximation property under inappropriate random parameter selection. 
In response to this issue, the stochastic configuration network (SCN) \cite{SCN} constrains random parameter selection by a novel supervisory mechanism to satisfy the universal approximation property with convergence in norm.
The supervisory mechanism of SCN clearly reveals that the assignment of uncertain random parameters \cite{Uncertain_Parameter} are not independent of sample data and should be constrained by data adaptively for better resolution of nonlinear optimization problems \cite{Linear_systems}. 
Therefore, the supervisory mechanism method is planned to be incorporated into BRLS, enhancing the feasibility of universal approximation property.

In this paper, we prove that the iterative errors of BRLS do not converge in norm, which implies that BRLS is infeasible if it is incrementally constructed with random parameters within a fixed range and its convergence rate meets certain conditions.
To enhance the feasibility of BRLS under arbitrary random parameter, we propose a supervisory mechanism with adaptive range settings into the BRLS network, thereby constructing the broad stochastic configuration residual learning system (BSCRLS) network. 
We mathematically demonstrate that the BSCRLS network possesses the universal approximation property with its iterative errors converging in norm manner.
In addition, three incremental BSCRLS networks adding enhancement nodes, feature nodes, and input data are present in response to the application requirements of various network updates. 
Solar panels dust detection experiments are performed on publicly available dataset SPDD \cite{kaggle} to demonstrate effectiveness of the proposed methods. Comparing with 13 relative deep and broad learning algorithms, experimental results show that BSCRLS outperforms other algorithms.
Notable contributions are as follows:
\begin{itemize}
  \item 
      We demonstrate that the iterative errors of BRLS do not converge in norm, indicating that the feasibility of BRLS's universal approximation property is limited.
  \item 
      We propose BSCRLS network based on a supervisory mechanism with adaptive range settings to enhance the feasibility of BRLS.
  \item 
      We prove the universal approximation property of the BSCRLS with its iterative errors converging in norm.
  \item 
      We present three incremental BSCRLS to satisfy the various incremental learning requirements, about increments of enhancement nodes, feature nodes, and input data.    
\end{itemize}

The remainder of this paper is organized as follows. 
In Section \ref{c2}, the related preliminary work about BRLS and SCN is described. 
In section \ref{c3}, we present Theorem \ref{counter-example} about the infeasibility of BRLS' universal approximation property in norm, and Theorem \ref{uniSC} about the universal approximation property in norm of the BSCRLS, as well as the algorithmic description of basic BSCRLS and three incremental BSCRLS networks.
In Section \ref{c4}, the experiments are conducted to demonstrate effectiveness of proposed methods. Finally, in Section \ref{c5} we draw the conclusion and further work.

In this paper, all norms are represented by $L_2$ norm. In SCN algorithm, for any Lebesgue measurable function $\boldsymbol{f}=[f_1,\dots,f_c]:\mathbb{R}^d \rightarrow \mathbb{R}^c$ on compact set $\boldsymbol{D}\subset\mathbb{R}^d$, $L_2$ norm is defined as
\begin{equation*}\label{SCN_L2Norm}
  \|\boldsymbol{f} \|:=(\sum_{r=1}^{c} \int_{\boldsymbol{D}}|f_r(x)|^2dx) ^\frac{1}{2}<\infty.
\end{equation*}
And the inner product of Lebesgue measurable functions $\boldsymbol{f}$ and $\boldsymbol{g}$ is  defined as
\begin{equation*}
  \langle\boldsymbol{f},\boldsymbol{g}\rangle:= \sum_{r=1}^{c} \boldsymbol{g} \langle f_r,g_r\rangle=\sum_{r=1}^{c} \int_{\boldsymbol{D}}f_r(x)g_r(x)dx.
\end{equation*}
where $\boldsymbol{g}\in L_2(\boldsymbol{D})$. 

In broad networks, for any Lebesgue measurable function in matrix form $\boldsymbol{F}=[f_{k,r}]_{N\times c}: \mathbb{R}^{N\times M} \rightarrow \mathbb{R}^{N\times c}$ on $\boldsymbol{D}'\subset\mathbb{R}^{N\times M}$, $L_2$ norm is defined as
\begin{equation*}
  \|\boldsymbol{F} \|_2:=(\sum_{k=1}^{N}\sum_{r=1}^{c} \int_{\boldsymbol{D}'}|f_{k,r}(x)|^2dx) ^\frac{1}{2}<\infty.
\end{equation*}

\section{Preliminaries}\label{c2}

In this section, we review BRLS and SCN methods.
BRLS algorithm constructs a sequence of residual operators with norm convergence to address the degradation in broad learning systems which could be viewed as an improved two-dimensional extension of the RVFL algorithm. 
However, the universal approximation property of BRLS algorithm is solely based on convergence in probability measure, similar to RVFL algorithm.
To address the limitation that the universal approximation property of RVFL algorithm is feasible only to a certain extent, 
SCN algorithm builds a supervisory mechanism.

\subsection{Broad residual learning system}

BRLS is a fast-dynamic-updating network to solve the degradation problem in broad networks by constructing a sequence of residual operators with norm convergence. The main components of BRLS networks are as follows.
\begin{enumerate}
  \item Generation of Feature nodes: Randomly assigning the input weights $\boldsymbol{W}_{e_i}$ and biases $\boldsymbol{\beta}_{e_i}$, then generating several groups of feature nodes according to corresponding stochastic parameters, $\boldsymbol{Z}_i=\phi_i(\boldsymbol{XW}_{e_i} + \boldsymbol{\beta}_{e_i})$, 
      where $\boldsymbol{X}$ is the training sample input 
      and $\phi _i,$ $i = 1,\dots, n$ are $n$ feature mapping functions. 
  \item Layer-1st residual learning: Generating layer-1st enhancement nodes $\boldsymbol{H}_1=\xi_1(\boldsymbol{Z}^n \boldsymbol{W}_{h_{1}} +\boldsymbol{\beta}_{h_{1}})$ based on 
      feature nodes $\boldsymbol{Z}^n=[\boldsymbol{Z}_1, \dots,\boldsymbol{Z}_n]$, activation function $\xi$ and randomly assigned input weights $\boldsymbol{W}_{h_{1}}$ and biases $\boldsymbol{\beta}_{h_{1}}$, then calculating the output weights of the 1st residual learning layer by ridge regression $\boldsymbol{W}_1 = (\boldsymbol{K}_1)^+ \boldsymbol{Y}$,
       where $\boldsymbol{K}_1=[\boldsymbol{Z}^n\vert \boldsymbol{H}_1]$ and $\boldsymbol{Y}$ is the sample labels.
  \item Layer-$j$th residual learning: Generating layer-$j$th enhancement nodes $\boldsymbol{K}_j=\xi_j (\boldsymbol{Z}^n\boldsymbol{W}_{h_{j}} +\boldsymbol{\beta}_{h_{j}}), j=2,\dots,m$ based on randomly assigned parameters $\boldsymbol{W}_{h_{j}}$ and $\boldsymbol{\beta}_{h_{j}}$, next determining the residual operator of current layer from the error between the sample labels and the output of first $j-1$ layers
      \begin{equation}\label{FXj-1}
      \boldsymbol{E}_{j}=
      \begin{cases}
      \boldsymbol{Y}-\boldsymbol{X}_{1} \boldsymbol{W}_1, & \mbox{if } j=1 \\
      \boldsymbol{E}_{j-1}- \boldsymbol{K}_{j} \boldsymbol{W}_j, & \mbox{if } j>1 
      \end{cases},
      \end{equation}
      then calculating the output weights of the $j$th residual learning layer by ridge regression $\boldsymbol{W}_j= (\boldsymbol{K}_{j})^+\boldsymbol{E}_{j-1}$.
  \item Determination the output weights of network: Collecting the output weights of all residual learning layers $\boldsymbol{W}^{(m)} = [(\boldsymbol{W}_1)^T,(\boldsymbol{W}_2)^T, \dots,(\boldsymbol{W}_m)^T]^T$. 
  \item Incremental learning: Generating new enhancement nodes based on different incremental learning requirements (adding enhancement nodes, feature nodes, and input data),
      next determining the corresponding residual operator of currant layer, 
      then calculating the output weights of this residual learning layer by ridge regression for connecting enhancement nodes to the residual operator. 
\end{enumerate}

The universal approximation theorem for any neural network algorithm is the fundamental basis. Our recent work demonstrates the universal approximation properties of BRLS by the convergence in probability measure in Theorem \ref{univer1}.

\begin{theorem}\label{univer1}
For any compact set $\boldsymbol{S}\subset\boldsymbol{I}^d$ and any vector-valued continuous function $\boldsymbol{f}=[f_1,f_2,\dots,f_c]$, $f_r\in C(\boldsymbol{I}^d)$, $r=1,\dots,c$, there exists a sequence of $\{\boldsymbol{f}_{\boldsymbol{\omega}_{n,m}}\}$ in BRLS  denoted as
$\boldsymbol{f}_{\boldsymbol{\omega}_{n,m}}= [f_{1,\boldsymbol{\omega}_{n,m}}, f_{2,\boldsymbol{\omega}_{n,m}},\dots, f_{c,\boldsymbol{\omega}_{n,m}}],$
such that
\begin{equation*}
  \lim_{n,m\rightarrow \infty} \rho_{\boldsymbol{S}} (\boldsymbol{f},\boldsymbol{f}_{\boldsymbol{\omega}_{n,m}})=0,
\end{equation*}
where $\rho_{\boldsymbol{S}}$ is the distance between $\boldsymbol{f}$ and $\boldsymbol{f}_{\boldsymbol{\omega}_{n,m}}$ on any compact set $\boldsymbol{S}\subset\boldsymbol{I}^d$ denoted as
\begin{equation*}\label{ps}
  \rho_{\boldsymbol{S}}(\boldsymbol{f}, \boldsymbol{f}_{\boldsymbol{\omega}_{n,m}})= (E\sum_{r=1}^{c} \int_{\boldsymbol{S}}[f_r(\boldsymbol{x})- f_{r,\boldsymbol{\omega}_{n,m}}(\boldsymbol{x})]^2 d\boldsymbol{x})^{\frac{1}{2}},
\end{equation*} 
and $E$ is the expectation with respect to $\mu_{n,m}$, as well as
\begin{equation*}\label{Brlsf}
\begin{split}
  f_{r,\boldsymbol{\omega}_{n,m}} (\boldsymbol{x})
  = &\sum_{i=1}^{nk} w_{r,z_i}\phi(\boldsymbol{xw}_{e_i} +\beta_{e_i})+ \\
  &\sum_{j=1}^{m} \sum_{l=1}^{q}w_{r,h_{j,l}}\xi(\boldsymbol{x};\{\phi, \boldsymbol{w}_{h_{j,l}}, \beta_{h_{j,l}}\})
\end{split}
\end{equation*}
is constructed by nonconstant bounded feature mapping function $\phi$, absolutely integrable activation function $\xi$ satisfying $\int_{\mathbb{R}^d}\xi^2 (\boldsymbol{x})d\boldsymbol{x}<\infty$, randomly generated parameters $\lambda_{n,m}=(\boldsymbol{w}_{e_1}, \dots,\boldsymbol{w}_{e_{nk}},\boldsymbol{w}_{h_{1,1}},\dots, \boldsymbol{w}_{h_{1,q}},\dots, \boldsymbol{w}_{h_{m,1}}, \dots, \boldsymbol{w}_{h_{m,q}},$ $\beta_{e_1},\dots, \beta_{e_{nk}},\beta_{h_{1,1}}, \dots, \beta_{h_{1,q}}, \dots, \beta_{h_{m,1}}, \dots, \beta_{h_{m,q}})$, and a respective sequence of probability measures $\mu_{n,m}$ to generate $\lambda_{n,m}$.
\end{theorem}

Building on Theorem \ref{univer1}, we present two corollaries that extend the domains of both activation and target functions: one expands the range of activation functions from square integrability to derivative with square integrability; and the other extends the range of approximable functions $\boldsymbol{f}$ from continuous to measurable functions. 
From these methods, it can be seen that the universal approximation theorem of BRLS is based on the convergence in probability measure. Probability measure is a likelihood measure under expectation and not strictly applicable to arbitrary cases relative to the norm measure.
In other words, the universal approximation property is conditional to BRLS networks and cannot be guaranteed in some special cases, which is justified in the next section (Theorem \ref{counter-example}). 

The similar issue also exists in RVFL algorithm, where its universal approximation property is only satisfied with convergence in probability measure and not in norm. Fortunately, SCN algorithm offers a solution to this issue.

\subsection{Stochastic configuration network}

SCN algorithm proposes a supervisory mechanism to constrain the generation of random parameters on the framework of RFVL algorithm, which makes its universal approximation property meet more rigorous norm convergence.

For any target function $\boldsymbol{f}=[f_1,\dots,f_c]:\mathbb{R}^d\rightarrow \mathbb{R}^c$, the SCN network with $m-1$ hidden nodes is set as $\boldsymbol{f}_{m-1}(\boldsymbol{x})=\sum_{j=1}^{m-1} \boldsymbol{\beta}_j \boldsymbol{g}_j(\boldsymbol{w}_j^T\boldsymbol{x} +\boldsymbol{b}_j)(m=1,2, \dots, f_0=0)$, where $\boldsymbol{\beta}_j=[\beta_{j,1},\dots, \beta_{j,c}]$. The current residual error is denoted as $\boldsymbol{e}_{m-1}=\boldsymbol{f} -\boldsymbol{f}_{m-1}=[e_{m-1,1},\dots,e_{m-1,c}]$.
If the error threshold is not satisfied, the new random basis function $\boldsymbol{g}_m$ is generated by random parameter $\boldsymbol{w}_m$ and $\boldsymbol{b}_m$ with the supervisory mechanism (inequality constraint).
The output weights $\boldsymbol{\beta}_m$ is evaluated by previous error $\boldsymbol{e}_{m-1}$ and new random basis function $\boldsymbol{g}_m$ to ensure the updated network $\boldsymbol{f}_m=\boldsymbol{f}_{m-1} +\boldsymbol{\beta}_m\boldsymbol{g}_m$ with an improved residual error. Theorem \ref{univer2} gives the detailed SCN construction theory and its universal approximation property converging in norm.

\begin{theorem}\label{univer2}
  Suppose that span($\Gamma$) is dense in $L_2$ space and $\forall \boldsymbol{g} \in \Gamma$, $0<\|\boldsymbol{g}\|< b_g$ for some $b_g\in\mathbb{R}^+$. Given
$0 < r < 1$ and a nonnegative real number sequence $\{\mu_m\}$ with $\lim_{m\to\infty}\mu_m=0$ and $\mu_m\leq(1-\gamma)$. For $m=1,2,\dots$, 
denote
\begin{equation}\label{scn_uni1}
  \delta_m=\sum_{r=1}^{c}\delta_{m,r},\ \delta_{m,r}=(1-\gamma-\mu_m) \|\boldsymbol{e}_{m-1,r}\|^2.
\end{equation}
If the random basis function $g_m$ is generated to satisfy the following inequalities:
\begin{equation}\label{scn_uni2}
  \langle\boldsymbol{e}_{m-1,r},\boldsymbol{g}_m \rangle^2\geq b_g^2\delta_{m,r},
\end{equation}
and the output weights are constructively evaluated by
\begin{equation}\label{scn_uni3}
  \beta_{m,r}=\frac{\langle\boldsymbol{e}_{m-1,r}, \boldsymbol{g}_m\rangle}{\|\boldsymbol{g}_m\|^2}
\end{equation}
Then, we have  $\lim_{m\to\infty}\|\boldsymbol{f} -\boldsymbol{f}_m\|=0$.
\end{theorem}

The supervisory mechanism by constraint conditions (\ref{scn_uni1}) and (\ref{scn_uni2}) restricts the arbitrary assignment of random parameters $\boldsymbol{w}_m$ and $\boldsymbol{b}_m$ in SCN, indicating that the configuration of random parameters should not be entirely reliant on the distribution and range settings, but is also related to the given training samples. 
The concept of constraining random parameters through the supervisory mechanism greatly inspires us in addressing the sensitivity of random parameter selection in BRLS. In the following, we adaptively propose a new supervisory mechanism within BRLS framework to constrain the selection of random parameters, ensuring that the universal approximation property of new algorithm converges in norm.


\section{Broad stochastic configuration residual learning system}\label{c3}

In this section, we demonstrate the infeasibility of BRLS networks for universal approximation under norm measure. Therefore, we propose a solution for constructing a novel BSCRLS network with a supervisory mechanism based on BRLS network. And the universal approximation property of BSCRLS is proved under more rigorous norm measure. Concretely, the basic BSCRLS and three incremental BSCRLS networks (adding enhancement nodes, feature nodes, and input data) are introduced.

\subsection{Limitations of universal approximation property in BRLS}

For a target function $\boldsymbol{F}:\mathbb{R}^{N\times M} \rightarrow \mathbb{R}^{N\times c}$, BRLS network starts from one residual layer 
$
  \boldsymbol{F}_1=\boldsymbol{K}_1\boldsymbol{W}_1,
$
where $\boldsymbol{K}_1$ is generated by  training samples and random parameters, and the output weight of this layer is calculated by $\boldsymbol{W}_1=(\boldsymbol{K}_1)^+\boldsymbol{E}_0$. Denote the initial error as
$
  \boldsymbol{E}_0=\boldsymbol{F}, 
$
and renew the error residual as $\boldsymbol{E}_1=\boldsymbol{F}-\boldsymbol{F}_1$. Then, BRLS network with $m$ residual learning layers is deterministic and described as
\begin{equation}\label{FL}
  \boldsymbol{F}_m=\boldsymbol{F}_{m-1}+\boldsymbol{K}_m \boldsymbol{W}_m,
\end{equation}
where $\boldsymbol{K}_m$ is generated by  training samples and random parameters, and the output weight of this layer is calculated by 
\begin{equation}\label{WL}
  \boldsymbol{W}_m=(\boldsymbol{K}_m)^+ \boldsymbol{E}_{m-1}.
\end{equation}
The current residual error is denoted as 
\begin{equation}\label{EL}
  \boldsymbol{E}_{m}=\boldsymbol{F}-\boldsymbol{F}_m. 
\end{equation}

\begin{theorem}\label{counter-example}
For BRLS network and sufficiently large $m$, if the following hold:
\begin{equation}\label{311}
  \frac{\|\boldsymbol{E}_{m-1}\|_2- \|\boldsymbol{E}_{m}\|_2}  {\|\boldsymbol{E}_{m-1}\|_2}\leq\varepsilon_m<1,
\end{equation}
and
\begin{equation}\label{312}
  \lim\limits_{m\to\infty}\prod_{j=1}^{m} (1-\varepsilon_j)=\varepsilon>0.
\end{equation}
Then, BRLS network has no universal approximation property, that is,
\begin{equation}\label{313}
  \lim\limits_{m\to\infty}\|\boldsymbol{F} -\boldsymbol{F}_m\|_2\geq\varepsilon \|\boldsymbol{F}\|_2.
\end{equation}
\end{theorem}

\emph{Proof:} 
According to Eq.(\ref{FL})-(\ref{EL}), we have  
\begin{equation*}
\begin{split}
   \|\boldsymbol{E}_{m}\|_2 & = \|\boldsymbol{F}-\boldsymbol{F}_{m}\|_2 \\
     & = \|\boldsymbol{F}-\boldsymbol{F}_{m-1} -\boldsymbol{K}_{m}\boldsymbol{W}_{m}\|_2\\
     & = \|\boldsymbol{E}_{m-1} -\boldsymbol{K}_{m}\boldsymbol{W}_{m}\|_2 \\
     & = \|\boldsymbol{E}_{m-1} -\boldsymbol{K}_{m}\boldsymbol{K}_{m}^+ \boldsymbol{E}_{m-1}\|_2 \\
     & = \|(\boldsymbol{I} -\boldsymbol{K}_{m}\boldsymbol{K}_{m}^+) \boldsymbol{E}_{m-1}\|_2 \\
     & \leq \|\boldsymbol{I} -\boldsymbol{K}_{m}\boldsymbol{K}_{m}^+\|_2  \|\boldsymbol{E}_{m-1}\|_2 .
\end{split}
\end{equation*}
For any matrix, considering that $L_2$ norm has unitary invariance and $rank(\boldsymbol{H}_{j+1})>0$ in general, $
  \|\boldsymbol{I} -\boldsymbol{K}_{m} \boldsymbol{K}_{m}^+\|_2 <1
$
is satisfied (Lemma 3.3 in \cite{BRLS}). So there is 
\begin{equation*}
  \|\boldsymbol{E}_{m}\|_2< \|\boldsymbol{E}_{m-1}\|_2,
\end{equation*}
that is, $\{\|\boldsymbol{E}_{m}\|_2\}$ is strictly monotonically decreasing. And $\{\|\boldsymbol{E}_{m}\|_2\}$ has a lower bound 0. According to the monotone convergence theorem, the norm sequence of residual operators $\{\|\boldsymbol{E}_{m}\|_2\}$ must converge.

However, if the decreasing rate of $\{\|\boldsymbol{E}_{m}\|_2\}$ meets conditions (\ref{311}) and (\ref{312}), it can be deduced that
\begin{equation*}
\begin{split}
   \|\boldsymbol{E}_{m}\|_2 & \geq (1-\varepsilon_m)\|\boldsymbol{E}_{m-1}\|_2 \\
     & \geq(1-\varepsilon_m)(1-\varepsilon_{m-1})\dots (1-\varepsilon_1)\|\boldsymbol{E}_{0}\|_2 \\
     & =\prod_{j=1}^{m} (1-\varepsilon_j)\|\boldsymbol{F}\|_2.
\end{split}
\end{equation*}
So we have
\begin{equation*}
  \lim\limits_{m\to\infty}\|\boldsymbol{E}_{m}\|_2\geq \lim\limits_{m\to\infty}\prod_{j=1}^{m} (1-\varepsilon_j)\|\boldsymbol{F}\|_2=\varepsilon \|\boldsymbol{F}\|_2.
\end{equation*}
$\hfill\square$

In BRLS network, $\boldsymbol{E}_{m}$ is convergent. But it is not rigorous enough for feasibility of the above incremental learning process, that is, if its decreasing rate is subject to constraint conditions (\ref{311}) and (\ref{312}), $\boldsymbol{E}_{m}$ is possibly not convergent to 0 and less than an expected tolerance $\varepsilon$ for sufficiently large $m$. In the following, we give a concrete example of a BRLS network satisfying conditions (\ref{311}) and (\ref{312}), which does not follow the universal approximation property.

\emph{Example:} 
For a nonnegative decreasing sequence $\{\varepsilon_m:\varepsilon_m=1/(4m^2)\}$, we get $\lim_{m\to\infty}\prod_{j=1}^{m} (1-\varepsilon_j)=2/\pi$. If BRLS network satisfies $(\|\boldsymbol{E}_{m-1}\|_2- \|\boldsymbol{E}_{m}\|_2)/  \|\boldsymbol{E}_{m-1}\|_2\leq1/(4m^2)$, BRLS network has no universal approximation property, that is, $\lim_{m\to\infty}\|\boldsymbol{f} -\boldsymbol{F}_m\|_2\geq(2/\pi) \|\boldsymbol{f}\|_2$.

Obviously, the universal approximation property is conditional to BRLS networks. It is crucial for BRLS networks to find the condition which can ensure the universal approximation property.

\subsection{Universal approximation property of BSCRLS}

In this subsection, we provide a constructive scheme which can consequently build a universal approximator, that is, broad stochastic configuration residual learning system. BSCRLS is an improved BRLS algorithm inspired by the concept of supervisory mechanism in SCN. In Theorem \ref{uniSC}, we give the detailed BSCRLS construction theory and its universal approximation property converging in norm.

\begin{theorem}\label{uniSC}
   Suppose that $m\in \mathbb{Z}_+$, $0<\gamma<1$ and a real number sequence $\{\mu_m: 0\leq\mu_m\leq(1-\gamma)\}$ with $\lim_{m\to\infty}\mu_m=0$. For BRLS network, if the inputs of each residual learning layer $\boldsymbol{K}_m$ which is generated by training samples and randomly parameters satisfies the inequality constraint:
  \begin{equation}\label{321}
    \|\boldsymbol{I}-\boldsymbol{K}_{m} (\boldsymbol{K}_{m})^+\| _2\leq(\gamma+\mu_m),
  \end{equation}
  then, we have $\lim_{m\to\infty}\|\boldsymbol{F} -\boldsymbol{F}_m\|_2=0$.
\end{theorem}

\emph{Proof.} In BRLS network, there is 
\begin{equation*}
  \| \boldsymbol{E}_{m}\|_2 
      \leq \|\boldsymbol{I} -\boldsymbol{K}_{m}\boldsymbol{K}_{m}^+\|_2  \|\boldsymbol{E}_{m-1}\|_2.
\end{equation*}
According to  the inequality constraint (\ref{321}), we have
\begin{equation*}
\begin{split}
   \|\boldsymbol{E}_{m}\|_2 
     & \leq (\gamma+\mu_m)\|\boldsymbol{E}_{m-1}\|_2 \\
     & = \gamma\|\boldsymbol{E}_{m-1}\|_2 +\mu_m\|\boldsymbol{E}_{m-1}\|_2.
\end{split}
\end{equation*}
Denote $\omega_m=\mu_m\|\boldsymbol{E}_{m-1}\|_2$. The above inequality is described as
\begin{equation*}
  \|\boldsymbol{E}_{m}\|_2 \leq \gamma \|\boldsymbol{E}_{m-1}\|_2+\omega_m.
\end{equation*}
Considering $\lim_{m\to\infty}\mu_m=0$, take the limit on both side of the inequality, and we obtain
\begin{equation*}
  \lim\limits_{m\to\infty}\|\boldsymbol{E}_{m}\|_2 \leq \gamma\lim\limits_{m\to\infty} \|\boldsymbol{E}_{m-1}\|_2.
\end{equation*}
Since $\{\|\boldsymbol{E}_{m}\|_2\}$ is convergent, $e=\lim_{m\to\infty}\|\boldsymbol{E}_{m}\|_2.$ 
Furthermore, we have $e\leq\gamma e$. Given that $0<\gamma<1$, it is clear that 
\begin{equation*}
  e=\lim_{m\to\infty}\|\boldsymbol{E}_{m}\|_2 =\lim_{m\to\infty}\|\boldsymbol{F} -\boldsymbol{F}_m\|_2=0.
\end{equation*}

$\hfill\square$

This algorithm design not only selects appropriate random parameters to generate newly added nodes $\boldsymbol{K}_{m}$ that meet the condition (\ref{321}), but also enforces the residual error to be zero along with the constructive process. Constrained random parameters make the configuration of random parameters not only depend on the distribution and parameter range, but also adapt to the information of the training sample. 
Thus, the supervision mechanism allows the BSCRLS to constrain excessive free randomness for optimization while considering the ability to map random parameters quickly and efficiently.

It is worth noting that the learning parameter $\gamma$ can be nonconstant and set as an increasing sequence approaching 1. This is based on the consideration that it becomes more challenging to find the appropriate parameters for newly added nodes as the residual error decreases. So the learning parameter $\gamma$ close to 1 makes BSCRLS algorithm more flexible to randomly search for suitable parameters in the case of smaller residuals.

\subsection{BSCRLS algorithm description}

\subsubsection{Basic BSCRLS}

Assume the training dataset with inputs $\boldsymbol{X}\in\mathbb{R}^{N\times M}$ and its corresponding labels $\boldsymbol{Y}\in\mathbb{R}^{N\times c}$. 
Based on input $\boldsymbol{X}$, $n$ groups of feature nodes 
$
  \boldsymbol{Z}_i=\phi_i(\boldsymbol{XW}_{e_i} + \boldsymbol{\beta}_{e_i}),\ i=1,\dots,n
$
and $m$ groups of enhancement nodes 
$
  \boldsymbol{H}_j=\xi_j(\boldsymbol{Z}^n\boldsymbol{W}_{h_{j}} +\boldsymbol{\beta}_{h_{j}}),\ j=1,\dots,m 
$
are generated, where the parameters $\boldsymbol{W}_{e_i}$, $\boldsymbol{W}_{h_{j}}$, $\boldsymbol{\beta}_{e_i}$ and $\boldsymbol{\beta}_{h_{j}}$ are randomly generated with proper dimensions, and 
$
  \boldsymbol{Z}^n=[\boldsymbol{Z}_1, \dots,\boldsymbol{Z}_n]
$
is the underlying variation factor. 
Denote $
     \boldsymbol{K}_{1} =[\boldsymbol{Z}^n\vert \boldsymbol{H}_1]$, 
      $\boldsymbol{K}_{j}  = \boldsymbol{H}_j$, $j=2,\dots,m$
as the inputs of each residual learning layer. 
Determine whether $\boldsymbol{K}_{j}$, $j=1,\dots,m$ satisfies Theorem \ref{uniSC}, and if not, regenerate until it does. For the BSCRLS network with one layer $\boldsymbol{F}_1=\boldsymbol{K}_1\boldsymbol{W}_1$, the output weight of this layer is calculated by 
\begin{equation}
  \boldsymbol{W}_1=(\boldsymbol{K}_1)^+\boldsymbol{E}_0,
\end{equation} where the initial error is $\boldsymbol{E}_0=\boldsymbol{Y}$. 
For the BSCRLS network with $m$ layer $\boldsymbol{F}_m=\boldsymbol{F}_{m-1}+\boldsymbol{K}_m \boldsymbol{W}_m$, the output weight of this layer is calculated by 
\begin{equation}
  \boldsymbol{W}_m=(\boldsymbol{K}_m)^+\boldsymbol{E}_{m-1},
\end{equation}
where the residual error is $\boldsymbol{E}_{m-1}=\boldsymbol{E}_{m-2}- \boldsymbol{K}_{m-1}\boldsymbol{W}_{m-1}$. 
The weights of all layers can be represented as 
\begin{equation}
  \boldsymbol{W}^{m} = [\boldsymbol{W}_1^T,\dots, \boldsymbol{W}_m^T]^T.
\end{equation}

\begin{algorithm*}[htp]
\caption{Basic BSCRLS}\label{algo1}
\begin{multicols}{2}
\begin{algorithmic}[1]
\STATE {\textbf{Input:}}  training samples
 $(\boldsymbol{X},\boldsymbol{Y})$;
 number of feature node groups $n$; number of enhancement node groups $m$; learning parameter $0<\gamma<1$; 
\FOR{$i=1;i\leq n$}
\STATE Random $\boldsymbol{W}_{e_i}$, $\boldsymbol{\beta}_{e_i}$;
\STATE Calculate $\boldsymbol{Z}_i=\phi_i(\boldsymbol{XW}_{e_i}+\boldsymbol{\beta}_{e_i})$;
\ENDFOR
\STATE Collect $\boldsymbol{Z}^n=[\boldsymbol{Z}_1,\dots,\boldsymbol{Z}_n]$;
\STATE Random $\boldsymbol{W}_{h_1}$, $\boldsymbol{\beta}_{h_1}$;
\STATE Calculate 
$\boldsymbol{H}_1=\xi_1(\boldsymbol{Z}^n\boldsymbol{W}_{h_{1}} +\boldsymbol{\beta}_{h_{1}})$;
\STATE Collect $\boldsymbol{K}_1=[\boldsymbol{Z}^n\vert \boldsymbol{H}_1]$;
\STATE Set $\mu_1=(1-\gamma)/2$;
\IF{$\boldsymbol{K}_1$ satisfies Theorem \ref{uniSC}}
    \STATE Break (go to \textbf{Step 16});
\ELSE
\STATE Return to \textbf{Step 7};
\ENDIF
\STATE Calculate $\boldsymbol{W}_1=(\boldsymbol{K}_1)^+\boldsymbol{E}_0$;
\FOR{$j=2;j\leq m$}
    \STATE Random $\boldsymbol{W}_{h_j}$, $\boldsymbol{\beta}_{h_j}$;
    \STATE Calculate $\boldsymbol{K}_j=\xi_j(\boldsymbol{Z}^n\boldsymbol{W}_{h_{j}}+\boldsymbol{\beta}_{h_{j}})$;
    \STATE Set $\mu_j=(1-\gamma)/(j+1)$;
    \IF{$\boldsymbol{K}_j$ satisfies Theorem \ref{uniSC}}
    \STATE Break (go to \textbf{Step 25});
    \ELSE
    \STATE Return to \textbf{Step 18};
    \ENDIF
    \STATE Calculate $\boldsymbol{E}_{j-1}=\boldsymbol{E}_{j-2}- \boldsymbol{K}_{j-1}\boldsymbol{W}_{j-1}$;
    \STATE Calculate $\boldsymbol{W}_{j}=(\boldsymbol{K}_j)^+ \boldsymbol{E}_{j-1}$;
\ENDFOR
\STATE 
 Collect $\boldsymbol{W}^{m} = [\boldsymbol{W}_1^T,\dots, \boldsymbol{W}_m^T]^T$;
\STATE {\textbf{Output:}}  $\boldsymbol{W}^{m}$.
\end{algorithmic}
\end{multicols}
\end{algorithm*}

\subsubsection{Incremental BSCRLS}

In response to the application requirements of various network updates, BSCRLS could make use of various incremental learning, including increments of enhancement nodes, feature nodes, and input data, to achieve fast network reconstruction.

\noindent\emph{(a) Incremental BSCRLS adding enhancement nodes}

When the network cannot achieve satisfactory results, BSCRLS can dynamically add enhancement nodes to train new residual learning layers for better learning ability.

Assume the newly-added enhancement nodes are 
\begin{equation}
  \boldsymbol{K}_{m+1}=\xi_{m+1}(\boldsymbol{Z}^n \boldsymbol{W}_{h_{m+1}}+\boldsymbol{\beta}_{h_{m+1}}),
\end{equation}
where $\boldsymbol{W}_{h_{m+1}}$ and $\boldsymbol{\beta}_{h_{m+1}}$ are randomly generated. 
Determine whether $\boldsymbol{K}_{m+1}$ satisfies Theorem \ref{uniSC}, and if not, regenerate until it does. 
The weights of $(m+1)$th residual learning layer can be calculated as
\begin{equation}
  \boldsymbol{W}_{m+1}=(\boldsymbol{K}_{m+1})^+ \boldsymbol{E}_{m}.
\end{equation}
So the weights of all layers is 
\begin{equation}
  \boldsymbol{W}^{m+1} = [(\boldsymbol{W}^{m})^T,\boldsymbol{W}_{m+1}^T]^T.
\end{equation}

\begin{algorithm*}[htp]
\caption{Incremental BSCRLS adding enhancement nodes}\label{algo2}
\begin{multicols}{2}
\begin{algorithmic}[1]
\STATE {\textbf{Input:}}  training samples
 $(\boldsymbol{X},\boldsymbol{Y})$;
 number of feature node groups $n$; number of enhancement node groups $m$; learning parameter $0<\gamma<1$; 
\FOR{$i=1;i\leq n$}
\STATE Random $\boldsymbol{W}_{e_i}$, $\boldsymbol{\beta}_{e_i}$;
\STATE Calculate $\boldsymbol{Z}_i=\phi_i(\boldsymbol{XW}_{e_i}+\boldsymbol{\beta}_{e_i})$;
\ENDFOR
\STATE Collect $\boldsymbol{Z}^n=[\boldsymbol{Z}_1,\dots,\boldsymbol{Z}_n]$;
\STATE Random $\boldsymbol{W}_{h_1}$, $\boldsymbol{\beta}_{h_1}$;
\STATE Calculate 
$\boldsymbol{H}_1=\xi_1(\boldsymbol{Z}^n\boldsymbol{W}_{h_{1}} +\boldsymbol{\beta}_{h_{1}})$;
\STATE Collect $\boldsymbol{K}_1=[\boldsymbol{Z}^n\vert \boldsymbol{H}_1]$;
\STATE Set $\mu_1=(1-\gamma)/2$;
\IF{$\boldsymbol{K}_1$ satisfies Theorem \ref{uniSC}}
    \STATE Break (go to \textbf{Step 16});
\ELSE
\STATE Return to \textbf{Step 7};
\ENDIF
\STATE Calculate $\boldsymbol{W}_1=(\boldsymbol{K}_1)^+\boldsymbol{E}_0$;
\FOR{$j=2;j\leq m$}
    \STATE Random $\boldsymbol{W}_{h_j}$, $\boldsymbol{\beta}_{h_j}$;
    \STATE Calculate $\boldsymbol{K}_j=\xi_j(\boldsymbol{Z}^n\boldsymbol{W}_{h_{j}}+\boldsymbol{\beta}_{h_{j}})$;
    \STATE Set $\mu_j=(1-\gamma)/(j+1)$;
    \IF{$\boldsymbol{K}_j$ satisfies Theorem \ref{uniSC}}
    \STATE Break (go to \textbf{Step 25});
    \ELSE
    \STATE Return to \textbf{Step 18};
    \ENDIF
    \STATE Calculate $\boldsymbol{E}_{j-1}=\boldsymbol{E}_{j-2}- \boldsymbol{K}_{j-1}\boldsymbol{W}_{j-1}$;
    \STATE Calculate $\boldsymbol{W}_{j}=(\boldsymbol{K}_j)^+ \boldsymbol{E}_{j-1}$;
\ENDFOR
\STATE 
 Collect $\boldsymbol{W}^{m} = [(\boldsymbol{W}_1)^T,\dots, (\boldsymbol{W}_m)^T]^T$;
\WHILE{The training error threshold is not satisfied}
    \STATE Random $\boldsymbol{W}_{h_{m+1}}$, $\boldsymbol{\beta}_{h_{m+1}}$;
    \STATE Calculate $\boldsymbol{K}_{m+1}=\xi_{m+1}(\boldsymbol{Z}^n\boldsymbol{W}_{h_{m+1}}+\boldsymbol{\beta}_{h_{m+1}})$;
    \STATE Set $\mu_{m+1}=(1-\gamma)/(m+2)$;
    \IF{$\boldsymbol{K}_{m+1}$ satisfies Theorem \ref{uniSC}}
    \STATE Break (go to \textbf{Step 11});
    \ELSE
    \STATE Return to \textbf{Step 3};
    \ENDIF
    \STATE Calculate $\boldsymbol{E}_{m}=\boldsymbol{E}_{m-1}- \boldsymbol{K}_{m}\boldsymbol{W}_{m}$;
    \STATE Calculate $\boldsymbol{W}_{m+1}=(\boldsymbol{K}_{m+1})^+ \boldsymbol{E}_{m}$;
    \STATE Collect $\boldsymbol{W}^{m+1} = [(\boldsymbol{W}^{m})^T,\boldsymbol{W}_{m+1}^T]^T$;
  \STATE Set $m=m+1$;
\ENDWHILE
\STATE {\textbf{Output:}}  $\boldsymbol{W}^{m}$.
\end{algorithmic}
\end{multicols}
\end{algorithm*}

\noindent \emph{(b) Incremental BSCRLS adding feature nodes}

Feature deficiency is an important factor affecting the model results, because of failed selection of enough underlying variation factors mapped to more discriminating feature space. BSCRLS can quickly add new feature nodes for more sufficient network representation. 

Assume that $(n+1)$th group of feature nodes are newly added and denoted by 
$
  \boldsymbol{Z}_{n+1}=\phi_{n+1}(\boldsymbol{XW}_{e_{n+1}} +\boldsymbol{\beta}_{e_{n+1}}),
$
where $\boldsymbol{W}_{e_{n+1}}$ and $\boldsymbol{\beta}_{e_{n+1}}$ are randomly generated. Naturally, all features can be collected as 
$
  \boldsymbol{Z}^{n+1}=[\boldsymbol{Z}^{n}, \boldsymbol{Z}_{n+1}].
$
Then, the input of the newly additional residual learning layer is denoted as 
\begin{equation}
  \boldsymbol{K}_{m+1}=[\boldsymbol{Z}_{n+1}\vert \xi_{m+1}(\boldsymbol{Z}^{n+1}\boldsymbol{W}_{h_{m+1}} +\boldsymbol{\beta}_{h_{m+1}})],
\end{equation}
where $\boldsymbol{W}_{h_{m+1}}$ and $\boldsymbol{\beta}_ {h_{m+1}}$ are randomly generated to calculate the enhancement nodes.
Determine whether $\boldsymbol{K}_{m+1}$ satisfies Theorem \ref{uniSC}, and if not, regenerate until it does. 
The weights of $(m+1)$th residual learning layer can be calculated as
\begin{equation}
  \boldsymbol{W}_{m+1}=(\boldsymbol{K}_{m+1})^+ \boldsymbol{E}_{m}.
\end{equation}

So the weights of all layers is 
\begin{equation}
  \boldsymbol{W}^{m+1} = [(\boldsymbol{W}^{m})^T,\boldsymbol{W}_{m+1}^T]^T.
\end{equation}

\begin{algorithm*}[htp]
\caption{Incremental BSCRLS adding feature nodes}\label{algo3}
\begin{multicols}{2}
\begin{algorithmic}[1]
\STATE {\textbf{Input:}}  training samples
 $(\boldsymbol{X},\boldsymbol{Y})$;
 number of feature node groups $n$; number of enhancement node groups $m$; learning parameter $0<\gamma<1$; 
\FOR{$i=1;i\leq n$}
\STATE Random $\boldsymbol{W}_{e_i}$, $\boldsymbol{\beta}_{e_i}$;
\STATE Calculate $\boldsymbol{Z}_i=\phi_i(\boldsymbol{XW}_{e_i}+\boldsymbol{\beta}_{e_i})$;
\ENDFOR
\STATE Collect $\boldsymbol{Z}^n=[\boldsymbol{Z}_1,\dots,\boldsymbol{Z}_n]$;
\STATE Random $\boldsymbol{W}_{h_1}$, $\boldsymbol{\beta}_{h_1}$;
\STATE Calculate 
$\boldsymbol{H}_1=\xi_1(\boldsymbol{Z}^n\boldsymbol{W}_{h_{1}} +\boldsymbol{\beta}_{h_{1}})$;
\STATE Collect $\boldsymbol{K}_1=[\boldsymbol{Z}^n\vert \boldsymbol{H}_1]$;
\STATE Set $\mu_1=(1-\gamma)/2$;
\IF{$\boldsymbol{K}_1$ satisfies Theorem \ref{uniSC}}
    \STATE Break (go to \textbf{Step 16});
\ELSE
\STATE Return to \textbf{Step 7};
\ENDIF
\STATE Calculate $\boldsymbol{W}_1=(\boldsymbol{K}_1)^+\boldsymbol{E}_0$;
\FOR{$j=2;j\leq m$}
    \STATE Random $\boldsymbol{W}_{h_j}$, $\boldsymbol{\beta}_{h_j}$;
    \STATE Calculate $\boldsymbol{K}_j=\xi_j(\boldsymbol{Z}^n\boldsymbol{W}_{h_{j}}+\boldsymbol{\beta}_{h_{j}})$;
    \STATE Set $\mu_j=(1-\gamma)/(j+1)$;
    \IF{$\boldsymbol{K}_j$ satisfies Theorem \ref{uniSC}}
    \STATE Break (go to \textbf{Step 25});
    \ELSE
    \STATE Return to \textbf{Step 18};
    \ENDIF
    \STATE Calculate $\boldsymbol{E}_{j-1}=\boldsymbol{E}_{j-2}- \boldsymbol{K}_{j-1}\boldsymbol{W}_{j-1}$;
    \STATE Calculate $\boldsymbol{W}_{j}=(\boldsymbol{K}_j)^+ \boldsymbol{E}_{j-1}$;
\ENDFOR
\STATE 
 Collect $\boldsymbol{W}^{m} = [(\boldsymbol{W}_1)^T,\dots, (\boldsymbol{W}_m)^T]^T$;
\WHILE{The training error threshold is not satisfied}
    \STATE Random $\boldsymbol{W}_{e_{n+1}}$, $\boldsymbol{\beta}_{e_{n+1}}$;
    \STATE Calculate $\boldsymbol{Z}_{n+1}=\phi_{n+1}(\boldsymbol{XW}_{e_{n+1}}+\boldsymbol{\beta}_{e_{n+1}})$;
    \STATE Collect $\boldsymbol{Z}^{n+1}=[\boldsymbol{Z}^{n},\boldsymbol{Z}_{n+1}]$;
    \STATE Random $\boldsymbol{W}_{h_{m+1}}$, $\boldsymbol{\beta}_{h_{m+1}}$;
    \STATE Calculate $\boldsymbol{K}_{m+1}=[\boldsymbol{Z}_{n+1}\vert \xi_{m+1}(\boldsymbol{Z}^{n+1}\boldsymbol{W}_{h_{m+1}} +\boldsymbol{\beta}_{h_{m+1}})]$;
    \STATE Set $\mu_{m+1}=(1-\gamma)/(m+2)$;
    \IF{$\boldsymbol{K}_{m+1}$ satisfies Theorem \ref{uniSC}}
    \STATE Break (go to \textbf{Step 14});
    \ELSE
    \STATE Return to \textbf{Step 6};
    \ENDIF
    \STATE Calculate $\boldsymbol{E}_{m}=\boldsymbol{E}_{m-1}- \boldsymbol{K}_{m}\boldsymbol{W}_{m}$;
    \STATE Calculate $\boldsymbol{W}_{m+1}=(\boldsymbol{K}_{m+1})^+ \boldsymbol{E}_{m}$;
    \STATE Collect $\boldsymbol{W}^{m+1} = [(\boldsymbol{W}^{m})^T,\boldsymbol{W}_{m+1}^T]^T$;
\STATE Set $n=n+1$, and $m=m+1$;
\ENDWHILE
\STATE {\textbf{Output:}}  $\boldsymbol{W}^{m}$.
\end{algorithmic}
\end{multicols}
\end{algorithm*}

\noindent \emph{(c) Incremental BSCRLS adding input data}
In practical applications, the amount of data has a tendency to gradually increase. In the face of newly added input data, an excellent learning method should have the ability to update trained model simply for new knowledge learning in the added data \cite{Increment2}. 
BSCRLS conveniently add the ever-increasing training samples into new residual learning layers for more effective network.

Assume that $\{(\boldsymbol{X_a},\boldsymbol{Y_a}):\boldsymbol{X_a}\in\mathbb{R}^{B\times M},\boldsymbol{Y_a}\in\mathbb{R}^{B\times c}\}$ is the new input data added to the model. The mapping feature of new input data is denoted as 
$
  \boldsymbol{Z}_{\boldsymbol{a}}^n=[\phi_1(\boldsymbol{X_aW}_{e_1} +\boldsymbol{\beta}_{e_1}),\dots, \phi_n(\boldsymbol{X_aW}_{e_n}+\boldsymbol{\beta}_{e_n})].
$
The corresponding generated enhancement nodes of new input data can be represented as 
$
  \boldsymbol{H}^m_{\boldsymbol{a}}=[\xi_1(\boldsymbol{Z}_{\boldsymbol{a}}^n \boldsymbol{W}_{h_{1}} +\boldsymbol{\beta}_{h_{1}}),\dots, \xi_m(\boldsymbol{Z}_{\boldsymbol{a}}^n \boldsymbol{W}_{h_{m}}+\boldsymbol{\beta}_{h_{m}})].
$
The feature and enhancement nodes of new input data can be combined as 
$
  \boldsymbol{X}_m^{\boldsymbol{a}}=[\boldsymbol{Z}_{\boldsymbol{a}}^n \vert \boldsymbol{H}^m_{\boldsymbol{a}}].
$
  The feature nodes of the total input data including the initial input data and the new input data are 
$
  \boldsymbol{Z}_A^{n}=[(\boldsymbol{Z}^n)^T, (\boldsymbol{Z}_{\boldsymbol{a}}^n)^T]^T.
$
In the newly $(m+1)$th residual learning layer, the input of newly additional residual learning layer is be denoted as 
\begin{equation}
  \boldsymbol{K}_{m+1}=\xi_{m+1}(\boldsymbol{Z}_A^{n} \boldsymbol{W}_{h_{m+1}}+\boldsymbol{\beta}_{h_{m+1}}),
\end{equation}
where $\boldsymbol{W}_{h_{m+1}}$ and $\boldsymbol{\beta}_{h_{m+1}}$ are randomly generated. 
Determine whether $\boldsymbol{K}_{m+1}$ satisfies Theorem \ref{uniSC}, and if not, regenerate until it does. 
The weights of $(m+1)$th residual learning layer can be calculated as
\begin{equation}
  \boldsymbol{W}_{m+1}=(\boldsymbol{K}_{m+1})^+ \boldsymbol{E}_{m},
\end{equation}
where the residual error is calculated by 
\begin{equation}
  \boldsymbol{E}_{m}
    =[(\boldsymbol{E}_{m-1}- \boldsymbol{K}_m\boldsymbol{W}_m)^T, (\boldsymbol{Y}_{\boldsymbol{a}} -\boldsymbol{X}_{m}^{\boldsymbol{a}} \boldsymbol{W}^{m})^T]^T.
\end{equation}
So the weights of all layers can be represented by 
\begin{equation}
  \boldsymbol{W}^{m+1} = [(\boldsymbol{W}^{m})^T,\boldsymbol{W}_{m+1}^T]^T.
\end{equation}

\begin{algorithm*}[htp]
\caption{Incremental BSCRLS adding input data}\label{algo4}
\begin{multicols}{2}
\begin{algorithmic}[1]
\STATE {\textbf{Input:}}  training samples
 $(\boldsymbol{X},\boldsymbol{Y})$;
 number of feature node groups $n$; number of enhancement node groups $m$; learning parameter $0<\gamma<1$; 
\FOR{$i=1;i\leq n$}
\STATE Random $\boldsymbol{W}_{e_i}$, $\boldsymbol{\beta}_{e_i}$;
\STATE Calculate $\boldsymbol{Z}_i=\phi_i(\boldsymbol{XW}_{e_i}+\boldsymbol{\beta}_{e_i})$;
\ENDFOR
\STATE Collect $\boldsymbol{Z}^n=[\boldsymbol{Z}_1,\dots,\boldsymbol{Z}_n]$;
\STATE Random $\boldsymbol{W}_{h_1}$, $\boldsymbol{\beta}_{h_1}$;
\STATE Calculate 
$\boldsymbol{H}_1=\xi_1(\boldsymbol{Z}^n\boldsymbol{W}_{h_{1}} +\boldsymbol{\beta}_{h_{1}})$;
\STATE Collect $\boldsymbol{K}_1=[\boldsymbol{Z}^n\vert \boldsymbol{H}_1]$;
\STATE Set $\mu_1=(1-\gamma)/2$;
\IF{$\boldsymbol{K}_1$ satisfies Theorem \ref{uniSC}}
    \STATE Break (go to \textbf{Step 16});
\ELSE
\STATE Return to \textbf{Step 7};
\ENDIF
\STATE Calculate $\boldsymbol{W}_1=(\boldsymbol{K}_1)^+\boldsymbol{E}_0$;
\FOR{$j=2;j\leq m$}
    \STATE Random $\boldsymbol{W}_{h_j}$, $\boldsymbol{\beta}_{h_j}$;
    \STATE Calculate $\boldsymbol{K}_j=\xi_j(\boldsymbol{Z}^n\boldsymbol{W}_{h_{j}}+\boldsymbol{\beta}_{h_{j}})$;
    \STATE Set $\mu_j=(1-\gamma)/(j+1)$;
    \IF{$\boldsymbol{K}_j$ satisfies Theorem \ref{uniSC}}
    \STATE Break (go to \textbf{Step 25});
    \ELSE
    \STATE Return to \textbf{Step 18};
    \ENDIF
    \STATE Calculate $\boldsymbol{E}_{j-1}=\boldsymbol{E}_{j-2}- \boldsymbol{K}_{j-1}\boldsymbol{W}_{j-1}$;
    \STATE Calculate $\boldsymbol{W}_{j}=(\boldsymbol{K}_j)^+ \boldsymbol{E}_{j-1}$;
\ENDFOR
\STATE 
 Collect $\boldsymbol{W}^{m} = [(\boldsymbol{W}_1)^T,\dots, (\boldsymbol{W}_m)^T]^T$;
\WHILE{The training error threshold is not satisfied}
\STATE {\textbf{Input:}} New samples
  $\{(\boldsymbol{X_a},\boldsymbol{Y_a})\}$ 
  are added;
    \STATE Calculate $\boldsymbol{Z}_{\boldsymbol{a}}^n=[\phi_1(\boldsymbol{X_aW}_{e_1} +\boldsymbol{\beta}_{e_1}),\dots,$ $ \phi_n(\boldsymbol{X_aW}_{e_n}+\boldsymbol{\beta}_{e_n})]$;
    \STATE Collect $\boldsymbol{Z}_A^{n}=[(\boldsymbol{Z}^n)^T, (\boldsymbol{Z}_{\boldsymbol{a}}^n)^T]^T$;
    \STATE Calculate $\boldsymbol{X}_m^{\boldsymbol{a}}=[\boldsymbol{Z}_{\boldsymbol{a}}^n \vert \xi_1(\boldsymbol{Z}_{\boldsymbol{a}}^n \boldsymbol{W}_{h_{1}} +\boldsymbol{\beta}_{h_{1}}),\dots,$ $  \xi_m(\boldsymbol{Z}_{\boldsymbol{a}}^n \boldsymbol{W}_{h_{m}}+\boldsymbol{\beta}_{h_{m}})]$;
    \STATE Random $\boldsymbol{W}_{h_{m+1}}$, $\boldsymbol{\beta}_{h_{m+1}}$;
    \STATE Calculate $\boldsymbol{K}_{m+1}=\xi_{m+1}(\boldsymbol{Z}^n_{A}\boldsymbol{W}_{h_{m+1}}+\boldsymbol{\beta}_{h_{m+1}})$;
    \STATE Calculate $\boldsymbol{E}_{m}
    =[(\boldsymbol{E}_{m-1}- \boldsymbol{K}_m\boldsymbol{W}_m)^T, (\boldsymbol{Y}_{\boldsymbol{a}} -\boldsymbol{X}_{m}^{\boldsymbol{a}} \boldsymbol{W}^{m})^T]^T$;
    \STATE Calculate $\boldsymbol{W}_{m+1}=(\boldsymbol{K}_{m+1})^+ \boldsymbol{E}_{m}$;    
    \STATE Collect $\boldsymbol{W}^{m+1} = [(\boldsymbol{W}^{m})^T,\boldsymbol{W}_{m+1}^T]^T$;
  \STATE Set $\boldsymbol{Z}^n=\boldsymbol{Z}_A^{n}$,  and $m=m+1$;
\ENDWHILE
\STATE {\textbf{Output:}}  $\boldsymbol{W}^{m}$.
\end{algorithmic}
\end{multicols}
\end{algorithm*}

\section{Experiments: Solar Panels Dust Detection }\label{c4}

In this section, performance of BSCRLS and incremental BSCRLS is evaluated by experiments of solar panels dust detection.
The experiments are implemented on a 64-bit operating system with 64GB RAM and Intel(R) Xeon(R) Gold 6248R CPU at 3.00GHz and 2.99GHz processor.

\subsection{Experimental introduction and dataset}
Solar panels, which function to convert sunlight into electrical energy, play an important role in solar photovoltaic systems. However, their efficiency and performance are influenced by environmental factors, particularly the accumulation of dust on their surfaces, which can substantially diminish solar energy conversion. In addition to reducing energy production, the buildup of dirt can affect the service life of solar panels by trapping moisture and causing corrosion, further compromising the panels' integrity and performance. Consequently, monitoring and timely cleaning solar panels is essential, necessitating an automated dust detection algorithm that can determine whether images of solar panels captured by drones are clean in order to increase modules efficiency, reduce maintenance cost and reducing the use of resources.

Our experiments are conducted on a publicly available dataset SPDD \cite{kaggle} from Kaggle, a renowned platform for data collections. The dataset consists of two kinds of RGB images labeled "Clean" and "Dusty" to investigate the ability of different machine learning classifiers to detect dust on solar panel surfaces with the highest possible accuracy. The dataset has 1493 "Clean" images and 1069 "Dusty" images with complex picture issues, including inconsistent picture sizes, noisy backgrounds, text overwrites, and potential mislabeling.

\begin{figure*}[h] 
\centering 
\includegraphics[scale=0.5]{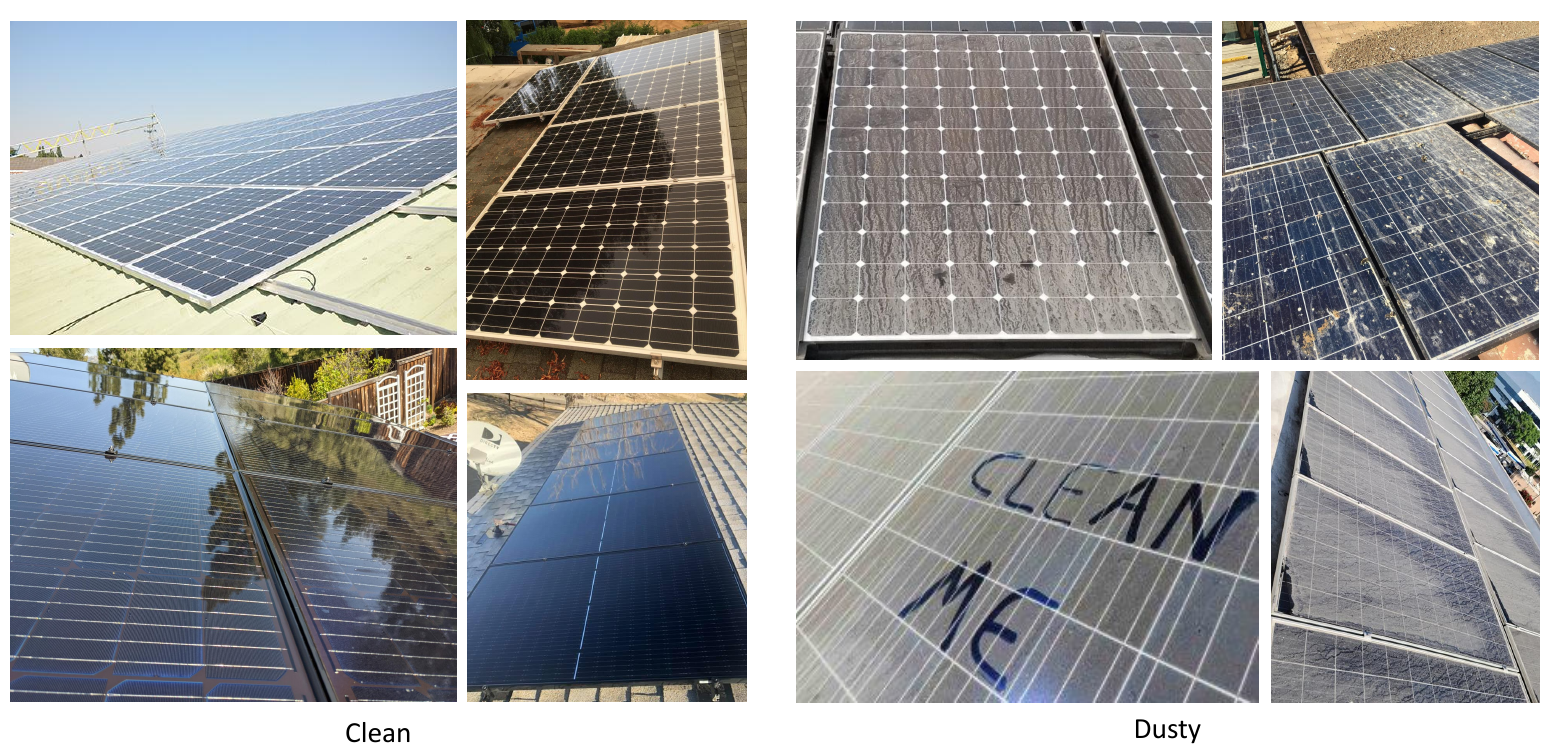}
\caption{Some sample images of solar panels dust detection}
\label{a2}
\end{figure*}

\begin{table}[htbp]
\centering
  \caption{Data augmentation during training to improve generalization and robustness}\label{b1}
\begin{tabular}{lc}
\toprule
\multicolumn{1}{c}{Augmentation metrics} & Parameter value \\ \midrule
Rotation range                           & 40              \\
Width shift range                        & 0.2             \\
Height shift range                       & 0.2             \\
Shear range                              & 0.2             \\
Zoom range                               & 0.2             \\
Horizontal flip                          &                 \\
Vertical flip                            &                 \\
Brightness range                         & 0.75-1.25       \\ 
\bottomrule
\end{tabular}
\end{table}

\begin{table*}[htp]
\renewcommand{\arraystretch}{0.5}
  \centering
  \caption{Performance of different algorithms on SPDD dataset}\label{e11}
\begin{tabular}{ccccccc}
\toprule
\textbf{Algorithm} & \textbf{Accuracy } & \textbf{ Precision} & \textbf{ Recall} & \textbf{ F1-score} & \textbf{False positive } & \textbf{ False negative}
\\
\midrule
SAE &  0.614$\pm$0.015  &  0.606$\pm$0.014  &  0.708$\pm$0.031  &  0.653$\pm$0.035  &  0.485$\pm$0.033  &  0.291$\pm$0.026
\\
SDA&  0.611$\pm$0.024  &  0.678$\pm$0.016  &  0.674$\pm$0.038  &  0.676$\pm$0.006  &  0.486$\pm$0.016  &  0.325$\pm$0.035 
\\
DBN&  0.634$\pm$0.028  &  0.652$\pm$0.017  &  0.713$\pm$0.018  &  0.682$\pm$0.015  &  0.462$\pm$0.014  &  0.286$\pm$0.038 
\\
DBM&   0.632$\pm$0.025  &  0.531$\pm$0.036  &  0.787$\pm$0.028  &  0.634$\pm$0.041  &  0.472$\pm$0.031  &  0.213$\pm$0.016  
\\
MLP&   0.591$\pm$0.023  &  0.586$\pm$0.020  &  0.687$\pm$0.018  &  0.632$\pm$0.011  &  0.508$\pm$0.006  &  0.313$\pm$0.021  
\\
FRBM& 0.586$\pm$0.019  &  0.550$\pm$0.036  &  0.697$\pm$0.037  &  0.615$\pm$0.013  &  0.512$\pm$0.039  &  0.302$\pm$0.016  
\\
AlexNet&    0.749$\pm$0.020  &  0.816$\pm$0.028  &  0.776$\pm$0.015  &  0.795$\pm$0.013  &  0.298$\pm$0.027  &  0.223$\pm$0.015   
\\
ResNet&   0.802$\pm$0.017  &  0.861$\pm$0.009  &  0.818$\pm$0.018  &  0.839$\pm$0.021  &  0.225$\pm$0.012  &  0.181$\pm$0.024   
\\
VGG&    0.788$\pm$0.034  &  0.877$\pm$0.018  &  0.791$\pm$0.031  &  0.832$\pm$0.035  &  0.220$\pm$0.012  &  0.208$\pm$0.009   
\\
MobileNet&   0.781$\pm$0.024 &  0.852$\pm$0.036  &  0.796$\pm$0.015  &  0.823$\pm$0.006  &  0.247$\pm$0.042  &  0.204$\pm$0.040   
\\
EfficientNet&   0.767$\pm$0.007  &  0.669$\pm$0.025  &  0.921$\pm$0.010  &  0.775$\pm$0.011  &  0.352$\pm$0.032  &  0.079$\pm$0.016   
\\
BLS&   0.794$\pm$0.033  &  0.821$\pm$0.018  &  0.834$\pm$0.030  &  0.827$\pm$0.008  &  0.262$\pm$0.029  &  0.165$\pm$0.025   
\\
BRLS&   0.864$\pm$0.008  &  0.796$\pm$0.030  &  0.970$\pm$0.017  &  0.875$\pm$0.014  &  0.240$\pm$0.036  &  0.029$\pm$0.007   
\\
\textbf{BSCRLS}&   \textbf{0.900$\pm$0.008}  &  \textbf{0.856$\pm$0.027} &  \textbf{0.974$\pm$0.017}  &  \textbf{0.911$\pm$0.033}  &  \textbf{0.182$\pm$0.021}  &  \textbf{0.026$\pm$0.024}  
\\
\bottomrule
\end{tabular}
\end{table*}

\begin{table}[h]
  \centering
  \caption{Training time of different algorithms on SPDD dataset}\label{e12}
\begin{tabular}{cc}
\toprule
\textbf{Algorithm}  & \textbf{Training Time (s)}\\
\midrule
SAE   & 6647.942  \\
SDA   & 9843.393 \\
DBN   & 15294.370 \\
DBM   & 23286.773  \\
MLP   & 7090.661 \\
FRBM   & 136.762 
\\
AlexNet&  4761.759
\\
ResNet&   20990.504
\\
VGG&  111756.763
\\
MobileNet& 2648.515  
\\
EfficientNet&  4290.226 
\\
BLS&  17.561 
\\
BRLS&   \textbf{6.794}
\\
\textbf{BSCRLS}&  8.172
\\
\bottomrule
\end{tabular}
\end{table}

\begin{table*}[htbp]
  \centering
  \caption{Comparative experiments between the incremental BRLS and BSCRLS in case of adding enhancement nodes on SPDD Dataset}\label{vs1}
\begin{tabular}{cccccccc}
\toprule
\multirow{2}{*}{\makecell{Feature \\Node Number}}  & \multirow{2}{*}{\makecell{Enhancement\\Node Number}} & \multicolumn{2}{c}{Accuracy } &
\multicolumn{2}{c}{\makecell{Additional Time (s)}} & \multicolumn{2}{c}{\makecell{Accumulative Time (s)}}\\
\cmidrule(lr){3-4} \cmidrule(lr){5-6} \cmidrule(lr){7-8}
& & BRLS & BSCRLS  & BRLS & BSCRLS  & BRLS & BSCRLS \\
\midrule
$10\times10$ & 100 & 0.506 & \textbf{0.522} & 0.128 & 0.178 & 0.128 & 0.178
\\
$10\times10$ & 100$\rightarrow$200 & 0.587 & \textbf{0.605} & 0.133 & 0.15 & 0.261 & 0.33 
\\
$10\times10$ & 200$\rightarrow$300 & 0.647 & \textbf{0.679} & 0.145 & 0.16 & 0.406 & 0.49
\\
$10\times10$ & 300$\rightarrow$400 & 0.705 & \textbf{0.732} & 0.135 & 0.172 & 0.541 & 0.662
\\
$10\times10$ & 400$\rightarrow$500 & 0.733 & \textbf{0.766} & 0.136 & 0.146 & 0.677 & 0.808
\\
$10\times10$ & 500$\rightarrow$600 & 0.740 & \textbf{0.782} & 0.152 & 0.178 & 0.829 & 0.986
\\
$10\times10$ & 600$\rightarrow$700 & 0.752 & \textbf{0.791} & 0.149 & 0.175 & 0.978 & 1.16
\\
$10\times10$ & 700$\rightarrow$800 & 0.765 & \textbf{0.809} & 0.137 & 0.155 & 1.115 & 1.316
\\
$10\times10$ & 800$\rightarrow$900 & 0.776 & \textbf{0.819} & 0.119 & 0.183 & 1.234 & 1.499
\\
$10\times10$ & 900$\rightarrow$1000 & 0.785 & \textbf{0.826} & 0.146 & 0.163 & 1.380 & 1.662
\\
\bottomrule
\end{tabular}
\end{table*}

\begin{table*}[htbp]
  \centering
  \caption{Comparative experiments between the incremental BRLS and BSCRLS in case of adding feature nodes on SPDD Dataset}\label{vs2}
\begin{tabular}{cccccccc}
\toprule
\multirow{2}{*}{\makecell{Feature \\Node Number}}  & \multirow{2}{*}{\makecell{Enhancement\\Node Number}} & \multicolumn{2}{c}{Accuracy } &
\multicolumn{2}{c}{\makecell{Additional Time (s)}} & \multicolumn{2}{c}{\makecell{Accumulative Time (s)}}\\
\cmidrule(lr){3-4} \cmidrule(lr){5-6} \cmidrule(lr){7-8}
& & BRLS & BSCRLS  & BRLS & BSCRLS  & BRLS & BSCRLS \\
\midrule
60 & 200 & 0.573 & \textbf{0.587} & 0.207 & 0.257 & 0.207 & 0.257
\\
60$\rightarrow$70 & 200$\rightarrow$400 & 0.668 & \textbf{0.711} & 0.209 & 0.26 & 0.416 & 0.517
\\
70$\rightarrow$80 & 400$\rightarrow$600 & 0.716 & \textbf{0.759} & 0.202 & 0.234 & 0.618 & 0.751
\\
80$\rightarrow$90 & 600$\rightarrow$800 & 0.740 & \textbf{0.773} & 0.224 & 0.274 & 0.842 & 1.025
\\
90$\rightarrow$100 & 800$\rightarrow$1000 & 0.762 & \textbf{0.791} & 0.198 & 0.256 & 1.040 & 1.281
\\
\bottomrule
\end{tabular}
\end{table*}

\begin{table*}[htbp]
  \centering
  \caption{Comparative experiments between the incremental BRLS and BSCRLS in case of adding input data on SPDD Dataset}\label{vs3}
\begin{tabular}{ccccccc}
\toprule
\multirow{2}{*}{\makecell{Input Data\\ Number}}  & \multicolumn{2}{c}{Accuracy } &
\multicolumn{2}{c}{\makecell{Additional Time (s)}} & \multicolumn{2}{c}{\makecell{Accumulative Time (s)}}\\
\cmidrule(lr){2-3} \cmidrule(lr){4-5} \cmidrule(lr){6-7}
& BRLS & BSCRLS  & BRLS & BSCRLS  & BRLS & BSCRLS \\
\midrule
1000 & 0.547 & \textbf{0.560} & 0.123 & 0.153 & 0.123 & 0.153
 \\
1000$\rightarrow$1100  & 0.649 & \textbf{0.692} & 0.136 & 0.172 & 0.259 & 0.325
\\
1100$\rightarrow$1200  & 0.703 & \textbf{0.735} & 0.140 & 0.199 & 0.399 & 0.524
\\
1200$\rightarrow$1300   & 0.736 & \textbf{0.759} & 0.153 & 0.206 & 0.552 & 0.730
\\
1300$\rightarrow$1400  & 0.751 & \textbf{0.774} & 0.169 & 0.213& 0.721 & 0.943
\\
\bottomrule
\end{tabular}
\end{table*}

\subsection{Data cleaning and preprocessing}

To enhance the dataset quality, 
we check the data labels and correct the faults. 
Considering the noisy backgrounds, the zero-shot object detection model, YOLO-World \cite{YOLO_World}, is utilized to crop images and save only the solar panel patches. The training/testing dataset division is 7:3. 
In the preprocessing phase, we normalize and augment the images to improve generalization and robustness. The data augmentation parameter is set as Table \ref{b1}.

\subsection{Experiment 1:  Performance comparison of BSCRLS with other algorithms on SPDD dataset}

In this subsection, the performance of basis BSCRLS on Algorithm \ref{algo1}is verified by comparing with other algorithms, including the stacked autoencoder (SAE) \cite{SAE}, the stacked denoised autoencoder (SDA) \cite{SDA}, the deep belief networks (DBN) \cite{DBN}, the deep Boltzmann machines (DBM)\cite{DBM}, the multilayer perceptron (MLP) \cite{MLP}, the fuzzy restricted Boltzmann machine (FRBM) \cite{FRBM}, AlexNet \cite{Alexnet}, ResNet \cite{ResNet}, 
MobileNet \cite{MobileNet}, EfficientNet \cite{EfficientNet}, BLS \cite{BLS}, and BRLS (2024, Under Review). 

Algorithms (SAE, SDA, DBN, DBM, MLP, FRBM, AlexNet, ResNet, VGG, MobileNet, and EfficientNet) are deep structures and the hyperparameters are tuned based on the back propagation. Their initial learning rates are set as 0.01 and the decay rate for each learning epoch is set as 0.95. The algorithms are trained for 50 epochs and a batch size of 16. The network structure of BLS, BRLS and BSCRLS about feature nodes and enhancement nodes is set as ($10\times10$,$100\times50$). The activation function is selected as sigmoid function \cite{Sigmoid}. The regularization parameter is set as $10^{-8}$. The associated random parameters about weights and biases are generated by standard uniform distributions on $[-1,1]$. The activation function, regularization parameter and parameter distribution of all BRLS and BLS networks in the later experiments will follow this setting and will not be detailed. Experiments are performed on SPDD dataset. 

Table \ref{e11} Table \ref{e12} and  show the comparative experimental results of these algorithms. It can be seen that BSCRLS has the best classification performance among other algorithms. In addition, BSCRLS algorithm is faster than other algorithms besides BRLS, due to the advantage of the fast speed of broad learning. The reason why BSCRLS is more time-consuming than BRLS is that BSCRLS is more rigorous in the decision process than BRLS. These experimental results fully verify the advantages of BSCRLS algorithm in accurate calculation and less time consumption.

\begin{figure*}[htbp] 
\centering 
\includegraphics[scale=0.85]{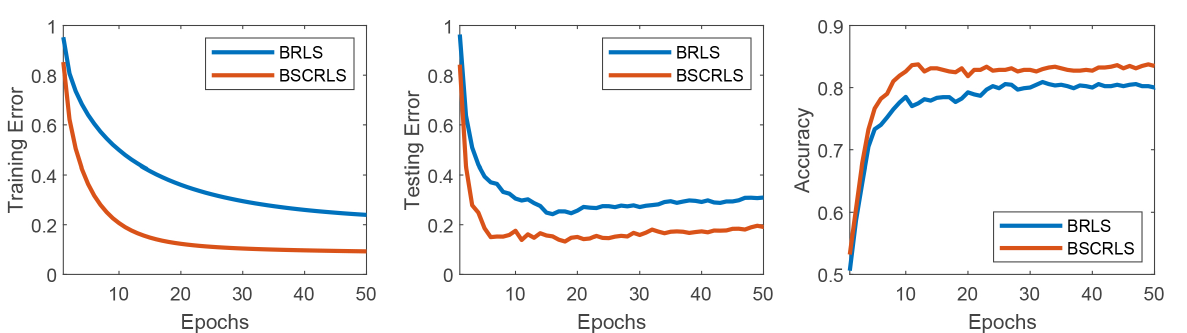}
\caption{Performance comparison of BSCRLS with convergence in norm and BRLS with convergence in probability measure}
\label{BrlsVsBscrls}
\end{figure*}

\subsection{Experiment 2: Performance comparison of BSCRLS with universal approximation property converging in norm and BRLS converging in probability measure}

In this subsection, we verify that the performance of BRLS with with universal approximation property converging in probability measure is weaker than BSCRLS with converging in norm. To reduce external factors, BRLS and BSCRLS adopt the same network structure with feature nodes and enhancement nodes as ($10\times10$,$100\times50$). Performance of two algorithms is reflected in training error, testing error, and testing accuracy. In training process, the experiment records the training errors of network layer 
\begin{equation*}
  Training\ Error(t)= \|\boldsymbol{E}_{t}\|_2.
\end{equation*}

In testing peocess, the testing error is 
\begin{equation*}
  Testing\ Error(t) =\|\boldsymbol{Y}'-\boldsymbol{X}_{t}' \boldsymbol{W}^{t}\|_2,
\end{equation*}
where $\boldsymbol{Y}'$ is the testing sample and $\boldsymbol{X}_{t}'$ is the testing feature and enhancement node matrix.
Accuracy is the percentage of the predicted correct number to the total number. 

The results are shown in Fig. \ref{BrlsVsBscrls}. We can easily see that the convergence of BSCRLS is stronger than BRLS. The accuracy of RSCRLS is also better than BRLS. These results fully demonstrate the superiority of BSCRLS with convergence in norm over BRLS with convergence in probability measure.

\subsection{Experiment 3: Effectiveness of incremental BSCRLS in three different incremental cases on SPDD dataset}

This subsection demonstrates the efficiency and the effectiveness of incremental BSCRLS adding enhancement nodes on Algorithm \ref{algo2}, feature nodes on Algorithm \ref{algo3}, and input data on Algorithm \ref{algo4}. 
The incremental BSCRLS and BRLS algorithms are compared under the same network settings on SPDD datasets. The advantages of three incremental BSCRLS are verified by comparative experiments with BRLS on classification accuracy and training time.

For the scenario about the increment of enhancement nodes, the initial BSCRLS network is set as $10\times10$ feature nodes and 100 enhancement nodes. Then, the incremental BSCRLS algorithm of Algorithm \ref{algo2} is applied to dynamically increase 100 enhancement nodes each layer until it reaches 1000. The incremental BRLS network for comparison has the same nodes. The experiment compares the accuracy and training time of the two incremental networks after each increment of enhancement nodes. The results are shown in Table \ref{vs1}. 
As can be seen from Table \ref{vs1}, the incremental BSCRLS network adding enhancement nodes could present more superior results on classification accuracy compared with BLS. However, the incremental BSCRLS network does not have the advantage of time consumption due to rigorous and complex decision process, while the incremental BRLS network performs better in training time because of the simplicity and efficiency of its incremental structure.

For the scenario about the increment of enhancement nodes, the BSCRLS network is initially set as $6\times10$ feature nodes and 200 enhancement nodes at the beginning of Algorithm \ref{algo3}. Then, the incremental BSCRLS network aims to dynamically increase feature nodes from $6\times10$ to $10\times10$ at the step of 10 in each increment. The corresponding enhancement nodes are increased at 200 each. As same as the incremental BRLS network structure. The each incremental experimental results are shown in Table \ref{vs2}. 
As can be seen from Table \ref{vs2}, the incremental BSCRLS network adding feature nodes could achieve more effective classification but slower training speed compared with the incremental BRLS network.

For the scenario about the increment of input data, the initial BSCRLS and BRLS networks can be trained with 1000 training samples. Then, incremental BRLS of Algorithm \ref{algo4} and corresponding
BRLS are applied to add dynamically 100 input data each time until they reach 1400. The incremental BSCRLS and BRLS networks are set as $10\times10$ feature nodes and $500$ enhancement nodes, where the corresponding updating enhancement nodes are increased at 100 each layer. Table \ref{vs3} presents the performance of the above two networks in the training times and classification results. As can be seen from Table \ref{vs3}, the incremental BSCRLS adding input data could complete classification tasks more efficiently.  

\section{Conclusion}\label{c5}

The BSCRLS methods proposed in this paper serve as a supplement to the BRLS methods, compensating for the insufficiency of BRLS's universal approximation property with  convergence solely in probability measure not in norm. 
By constructing a supervisory mechanism with conditional and adaptive allocation of random parameters, BSCRLS has the strict universal approximation property with convergence in norm, and surmounts the infeasibility issues of BRLS under improper random parameter allocation. 
BSCRLS has constructed a more rigorous and comprehensive theoretical framework, which will lay a solid foundation for the further development of broad learning algorithms.

Although BSCRLS has a good performance in solar
panels dust detection task, it needs to be verified in many research fields as a new algorithm. Therefore, our next work will focus on exploring more practical applications of BSCRLS for machine learning tasks.

\bibliographystyle{cas-model2-names}

\bibliography{bibliography}


\end{document}